\def\year{2021}\relax
\documentclass[letterpaper]{article} 
\usepackage{mystyle}  
\usepackage{times}  
\usepackage{helvet} 
\usepackage{courier}  
\usepackage[hyphens]{url}  
\usepackage{graphicx} 
\usepackage{graphicx}  
\usepackage{natbib}  
\usepackage{caption} 
\nocopyright
\usepackage{amsmath,amsfonts}
\usepackage{algorithm}
\usepackage[noend]{algpseudocode}

\newtheorem{theorem}{Theorem}

\title{Estimation of Structural Causal Model \\
    via Sparsely Mixing Independent Component Analysis}

\author{
    Kazuharu Harada,\textsuperscript{\rm 1}
    Hironori Fujisawa\textsuperscript{\rm 2, 1, 3} \\
}
\affiliations {
    \textsuperscript{\rm 1} 
        The Graduate University for Advanced Studies (SOKENDAI), Tokyo \\
    \textsuperscript{\rm 2} 
        The Institute of Statistical Mathematics, Tokyo \\
    \textsuperscript{\rm 3} 
        RIKEN Center for Advanced Intelligence Project (AIP), Tokyo \\
    \{kharada, fujisawa\}@ism.ac.jp
}

\begin{document}
\maketitle
\begin{abstract}
We consider the problem of inferring the causal structure from observational data, especially when the structure is sparse. This type of problem is usually formulated as an inference of a directed acyclic graph (DAG) model. The linear non-Gaussian acyclic model (LiNGAM) is one of the most successful DAG models, and various estimation methods have been developed. However, existing methods are not efficient for some reasons: (i) the sparse structure is not always incorporated in causal order estimation, and (ii) the whole information of the data is not used in parameter estimation. To address {these issues}, we propose a new estimation method for a linear DAG model with non-Gaussian noises. The proposed method is based on the log-likelihood of independent component analysis (ICA) with two penalty terms related to the sparsity and the consistency condition. The proposed method enables us to estimate the causal order and the parameters simultaneously.
For stable and efficient optimization, we propose some devices, such as a modified natural gradient. Numerical experiments show that the proposed method outperforms existing methods, including LiNGAM and NOTEARS.
\end{abstract}

\noindent 
It has been a significant topic to search for causal relationships from observational data in various scientific fields, such as biology \cite{sachs2005causal}, economy \cite{hoover2008causality}, and psychology \cite{glymour1998learning}. 
To detect causal relationships, graphical models have been used, in which directed edges represent causal relationships. In particular, a Directed Acyclic Graph (DAG) has been intensively studied for years. There are some traditional approaches for learning a DAG, like PC algorithm \cite{spirtes2000causation} and GES algorithm \cite{chickering2002optimal}. {These} can produce a Markov equivalence class, which contains all DAGs consistent with the conditional independence of the data, but cannot always identify a complete causal structure \cite{spirtes2000causation,judea2000causality} . 

This problem has been overcome by \citet{shimizu2006linear}, \citet{zhang2009identifiability}, \citet{peters2014identifiability}, and \citet{peters2014causal}. They constructed fully-identifiable DAG models by some additional assumptions. In particular, the Linear Non-Gaussian Acyclic Model (LiNGAM) proposed by \citet{shimizu2006linear} has many applications in various fields \citep[e.g.][]{liu2015altered, von2012directional,xu2017contemporaneous}, {by virtue of} good interpretability and relatively mild assumptions, compared to the other identifiable linear DAG models.

\citet{shimizu2006linear} also proposed an estimation method based on Independent Component Analysis (ICA). {Recently,} some other estimation methods for LiNGAM were proposed: DirectLiNGAM \cite{shimizu2011directlingam}, Pairwise LiNGAM \cite{hyvarinen2013pairwise}, and High-dimensional LiNGAM \cite{wang2020high}. 
These methods, including the original LiNGAM, consist of two stages, causal order estimation and parameter estimation. 

Here, we focus on a sparse structure of causal relationships in high-dimensional data. The two-stage methods are not efficient in this case for two reasons: (i) {the} sparse structure is not always incorporated in causal order estimation, and (ii) the whole information of the data is not used in parameter estimation. 
To address these issues, we propose a new one-stage method based on LiNGAM with sparsity. Based on the whole information of the data, the method estimates the causal order and the parameters simultaneously. A sparse estimation method based on LiNGAM was already discussed by \citet{zhang2006ica} and \citet{zhang2009ica}. However, their methods do not satisfy the consistency conditions of the optimal solution. This paper addresses this issue by a devised penalty related to the consistency condition. These issues motivate us to develop a one-stage algorithm incorporating the sparse structure.

In this paper, we formulate a one-stage estimation method for LiNGAM with a sparse penalty and a devised penalty related to the consistency condition of the optimal solution. 
Then, we make an efficient optimization algorithm that modifies a widely-used ADMM algorithm with a devised expression and {a modified} natural gradient descent.
We also propose an objective procedure for selecting a tuning parameter via the likelihood cross validation (CV). 
{In addition}, we demonstrate the effectiveness and scalability of the proposed method through numerical experiments. The proposed method is compared with some estimation methods of LiNGAM and another approach NOTEARS \cite{zheng2018dags}. The proposed method outperforms others in almost all cases and shows stable performance even in high dimensional cases. The scalability of the proposed method is examined. The proposed method is also applied to real data.

\section{Problem and Related Works}

\subsection{Linear DAG Model}

Suppose that $X=(X_1,\ldots,X_d)^T$ is a $d$-dimensional random vector drawn from the joint distribution $P(X)$. We assume ${\rm E}[X]=0$ without loss of generality. A Directed Acyclic Graph (DAG) model on $X$ can be defined by $P(X)$ and a DAG $G=(V_G, E_G)$, where $V_G$ and  $E_G$ are the sets of vertices and directed edges, respectively. The vertices in $V_G$ correspond to the indices of $X$. The element $(j,k) \in E_G$ shows a directed edge from $j$ to $k$, which represents the direct causal relationship from $X_j$ to $X_k$. The acyclicity implies no directed cycles in $G$.

This paper focuses on a linear DAG model.
A linear DAG model is equivalent to a linear structural equation given by
\begin{align}
    X_j &= \mathbf{b}_j^T X + S_j, \qquad j=1,\ldots,d,
\end{align}
where $\mathbf{b}_j=(b_{j1},\ldots,b_{jd})^T$ is a coefficient parameter vector and $S=(S_1,\ldots,S_d)^T$ is {an} independent noise vector. Suppose that $(j,k)\not\in E_G$ if $b_{kj}=0$ and $(j,k)\in E_G$ if $b_{kj}\ne 0$, which defines the structure of $G$. 
The above equation can also be expressed in matrix form:
\begin{equation} \label{eq:SEM}
    X = \mathbf{B}X + S,
\end{equation}
where $\mathbf{B}=(\mathbf{b}_1,\ldots,\mathbf{b}_d)^T$ is called the weighted adjacency matrix in graph theory.
From the acyclicity, all the diagonal entries of $\mathbf{B}$ are zero and $X_1,\ldots,X_d$ can be rearranged in {a} {\it causal order}. Let $\pi(j)$ be the index of $X_j$ permuted in {a} causal order, and then we have $\pi(j)<\pi(k)$ for any edges $(\pi(j),\pi(k))\in E_G$.
The corresponding permutation matrix $\mathbf{P}$ can transform $\mathbf{B}$ by $\mathbf{PBP}^T$ to a strictly lower triangular matrix \cite{shimizu2006linear,bollen1989}. 

Suppose that we have the data $\mathbf{X}\in\mathbb{R}^{N\times d}$ with $N$ random samples drawn from $P(X)$. The problem is to estimate the parameter matrix $\mathbf{B}$ from the data $\mathbf{X}$. 
The difficulty in the problem is {that} $\mathbf{B}$ must be transformed by a permutation to a lower triangular matrix due to the acyclicity.

\subsection{LiNGAM and Its Estimation}
    LiNGAM \cite{shimizu2006linear} is a linear DAG model with non-Gaussian noises, which is proved to be fully identifiable via Independent Component Analysis (ICA).
    The linear DAG model can be expressed by 
    \begin{align}
        X = \mathbf{M}^{-1}S, \qquad \mathbf{M}=\mathbf{I}-\mathbf{B},
    \end{align}
    which can be seen as an ICA model with the mixture matrix $\mathbf{M}^{-1}$. In the ICA model, the matrix $\mathbf{M}$ is identifiable up to the scaling and the permutation of its rows. 
    This indeterminacy is not a problem in ICA, but a crucial problem in causal discovery. This problem can be solved using {the} acyclicity of $G$, as described in \citet{shimizu2006linear}.

    \citet{shimizu2006linear} also constructed the parameter estimation algorithm, ICA-LiNGAM. This algorithm is summarized as follows. First, the parameter estimate $\hat{\mathbf{M}}_{ICA}$ is obtained by a standard ICA algorithm like FastICA \cite{hyvarinen1999fast}. 
    Second, the rows of $\hat{\mathbf{M}}_{ICA}$ are permuted so that its diagonal entries are far from zero as possible, and then each row is rescaled by its corresponding diagonal entry. 
    Let $\hat{\mathbf{M}}$ be the permuted and rescaled matrix. 
    Third, the causal order is explored by minimizing the squared sum of upper-triangular entries of $\hat{\mathbf{B}}~(= \mathbf{I}-\hat{\mathbf{M}})$. 
    Here, there is a big problem that the third step is computationally infeasible even when $d$ is not large.
    In order to avoid this problem, non-zero entries with small absolute values in $\hat{\mathbf{B}}$ are sequentially truncated to zero until $\hat{\mathbf{B}}$ can be transformed into a lower triangular matrix.
    After the transformation, $X$ is rearranged in a causal order. Finally the parameter is estimated again by sparse linear regression along with the determined ordering \cite{wiedermann2016statistics, zou2006adaptive}.
    
    \citet{zhang2006ica} and \citet{zhang2009ica} proposed another approach based on the log-likelihood of ICA with a sparse penalty. However, it does not satisfy the consistency {conditions} of independent components, as discussed later, so that the parameter estimation {is considered to be} unstable. In addition, the subgradient of the sparse penalty is approximated by a continuous function, and thereby the estimate is not always shrunk to zero. 
    This paper adopts this line, but develops the penalty term to meet the consistency assumptions and essentially improves the parameter estimation algorithm to make it stable and obtain exact sparsity.
    
    A different approach from ICA-based methods {was} employed in DirectLiNGAM \cite{shimizu2011directlingam}, Pairwise LiNGAM \cite{hyvarinen2013pairwise}, and High-dimensional LiNGAM \cite{wang2020high}. 
    These methods evaluate the causal order of variable pairs and then integrate them to obtain a global causal order. Next, they estimate the parameter in the same manner as that of ICA-LiNGAM.
    These approaches are preferable because the solution can be obtained in finite steps in contrast to the iterative algorithms.
    However, there are two difficulties in this line.
    The first one is that the causal order estimation is not efficient because the sparse structure of $\mathbf{B}$ is not incorporated, except for \citet{wang2020high}.
    The second one is that the parameter estimation of this approach is based on the least squares. This {means} the assumptions of non-Gaussianity is not used in parameter estimation, so that the efficiency may be reduced due to the information loss.
    
    \subsection{NOTEARS}
    \citet{zheng2018dags} introduced a new characterization of "DAGness" using a continuous function $h: \mathbb{R}^{d\times d}\rightarrow\mathbb{R}$ such that $h(\mathbf{B})=0$ if and only if the graph is acyclic. 
    {The function $h$} enables us to obtain DAG via totally continuous optimization, although the structure of DAG was explored via combinatorial optimization in the past.
    Using {the} function $h$, they proposed a novel structure learning method by the following continuous optimization problem:
    \begin{align}\label{eq:notears}
        \begin{array}{c}\displaystyle 
             \min_{\mathbf{B}\in\mathbb{R}^{d\times d}} \quad {\frac{1}{2N}}\|\mathbf{X}-\mathbf{XB}^T\|_F^2 + \lambda\|\mathbf{B}\|_1\\
             \text{subject to} \quad h(\mathbf{B})=0.
        \end{array} 
    \end{align}
    This {was} called Non-combinatorial Optimization via Trace Exponential and Augmented lagRangian for Structure learning (NOTEARS).
    
    NOTEARS does not assume any conditions on noises except for independence. However, we suspect this assumption because the linear DAG model is not identifiable without some additional conditions \cite{spirtes2000causation}; e.g., if the noises are assumed to be Gaussian with \textit{equal variances} and the squared error in \eqref{eq:notears} is replaced by the negative log-likelihood of Gaussian noises, the linear DAG model is identifiable \cite{peters2014identifiability}. This assumption, in particular, the assumption of equal variances, will be much stronger in real-world data than the non-Gaussian assumption in LiNGAM. 


\section{Parameter Estimation}

In this section, we propose a new method, which we call sparsely mixing ICA-LiNGAM (sICA-LiNGAM), for estimating the linear DAG model with non-Gaussian noises. The parameter is estimated using the log-likelihood of ICA with two penalties. Although a usual ICA method needs the orthogonality of the parameter matrix, it will not work well with a sparse penalty, as described later. To address this issue, we introduced a devised penalty.

\subsection{ICA and Consistent Estimation}
Here we briefly review the ICA and its consistent estimation. Let $S=(S_1,\ldots,S_d)^T$ be the vector of independent components (ICs) with zero means. Suppose that we observe $X={\mathbf{A}}S$. The purpose of ICA is to recover the ICs $S$ by ${\mathbf{M}}X$, where $\mathbf{M}$ is the parameter matrix.

Let $p_j$ be the probability density function of $S_j$ and let $\mathbf{M}=(\boldsymbol{m}_1,\ldots,\boldsymbol{m}_d)^T$ and $\mathbf{X}=(\boldsymbol{x}_1,\ldots,\boldsymbol{x}_N)^T$.
The maximum likelihood estimator (MLE) $\hat{\mathbf{M}}$ is given by 
\begin{align} \label{eq:likM}
    \hat{\mathbf{M}}=& \arg\max_{\mathbf{M}}\ell(\mathbf{M}; \mathbf{X}), \\
    \ell(\mathbf{M}; \mathbf{X}) 
        =& \frac{1}{N}\sum_{i=1}^{N}\sum_{j=1}^d \log p_j(\boldsymbol{m}_j^T\boldsymbol{x}_i)
            + \log|\det\mathbf{M}|.
\end{align}

In ICA, we assume that $p_j$s are unknown, so that it seems impossible to estimate $\mathbf{M}$ consistently. However, even when $p_j$s are unknown, the following theorem tells that MLE has consistency except for the indeterminacy of the scale and the order of the estimated ICs if we use appropriate probability density functions $\tilde{p}_j$s instead of $p_j$s. 

\begin{theorem}[\citet{hyvarinen2001independent}] \label{thm1}
    Let $Y_j=\boldsymbol{m}_j^T X$ be the estimated ICs. Suppose that $Y_j$s are uncorrelated with unit variance. The MLE $\hat{\mathbf{M}}$ has consistency except for the indeterminacy of the scale and the order of ICs, if 
    \begin{align} \label{eq:nlmoment}
        \mathbb{E}[S_j(\tilde{g}_j(S_j)) - \tilde{g}_j'(S_j)] > 0 \quad \text{for }  j=1,\ldots,d, 
    \end{align}
where
    \begin{align} \label{eq:gfunc}
        \tilde{g}_j(S_j)=\frac{\partial}{\partial S_j}\log \tilde{p}_j(S_j).
    \end{align}
\end{theorem}

Pre-whitening is often employed in ICA algorithms to convert the assumptions on $Y_j$s in Theorem \ref{thm1} into a convenient parameter constraint. 
Consider the spectral decomposition of $(1/N)\mathbf{X}^T\mathbf{X}=\mathbf{VD}^2\mathbf{V}^T$, where $\mathbf{V}$ is the orthogonal matrix and $\mathbf{D}^2$ is the diagonal matrix whose diagonal entries are the eigenvalues of $(1/N)\mathbf{X}^T\mathbf{X}$.
Let
\begin{align}
    \mathbf{Z}
        = (\mathbf{z}_1,\ldots,\mathbf{z}_N)^T
        = \mathbf{X}  \mathbf{V} \mathbf{D}^{-1},
\end{align}
where the diagonal elements of $\mathbf{D}$ are non-negative.
Then we have $(1/N)\mathbf{Z}^T\mathbf{Z}=I_d$, so that
the transformed variable $\mathbf{Z}$ is uncorrelated {with unit variance} before estimation. Let $\mathbf{W}=\mathbf{MVD}$. 
Therefore, the MLE of $\mathbf{W}$ is given by
\begin{align} 
    \hat{\mathbf{W}}=& \arg\max_{\mathbf{W}}\tilde\ell(\mathbf{W}; \mathbf{Z}), \\
    \tilde\ell(\mathbf{W}; \mathbf{Z}) 
        =& \frac{1}{N}\sum_{i=1}^{N}\sum_{j=1}^d \log p_j(\boldsymbol{w}_j^T\boldsymbol{z}_i)
            + \log|\det\mathbf{W}|, \label{eq:likW}
\end{align}
where $\boldsymbol{w}_j$ is the $j$th row of $\mathbf{W}$.
We also have $\hat{\mathbf{M}}=\hat{\mathbf{W}}\mathbf{D}^{-1}\mathbf{V}^T$.

Let {$Y_{ij}=\boldsymbol{m}^T_j\boldsymbol{x}_i$} and  $\mathbf{Y}=(Y_{ij})\in\mathbb{R}^{N\times d}$. We have $\mathbf{Y}=\mathbf{XM}^T=\mathbf{ZW}^T$.
Here, we suppose $\mathbf{W}$ is an orthogonal matrix. Considering the pre-whitening, the orthogonality of $\mathbf{W}$ keeps $\mathbf{Y}$ whitened:
\begin{align} \label{eq:ortho}
    \frac{1}{N}\mathbf{Y}^T\mathbf{Y}
        = \mathbf{W}\left(\frac{1}{N}\mathbf{Z}^T\mathbf{Z}\right)\mathbf{W}^T
        = \mathbf{W}\mathbf{W}^T = \mathbf{I}.
\end{align}
This implies that if we assume $\mathbf{W}$ is an orthogonal matrix, the estimated ICs are whitened even in a finite sample. Many ICA algorithms efficiently obtain $\hat{\mathbf{W}}$ by utilizing this orthogonality \cite{hyvarinen2001independent}. 
However, \eqref{eq:ortho} should hold at the population level, as seen in Theorem \ref{thm1}, and hence the orthogonality constraint is not necessary to be satisfied
strictly in a finite sample. 
In addition, a strict orthogonality constraint will be difficult to meet a sparse estimation, as mentioned later, and therefore
we relax this constraint as a penalty term, as described later.

\subsection{ICA with Sparse Penalty for Causal Discovery}

Returning to the linear DAG estimation, we consider the sparse structure of $\mathbf{M}$, because the half of the entries of $\mathbf{M}~(=\mathbf{I}-\mathbf{B})$ should be zero due to the acyclicity. In addition, the causal structure may be simple, which also implies $\mathbf{M}$ has a sparse structure.
We add to the log-likelihood \eqref{eq:likW} a sparse penalty of the adaptive lasso \cite{zou2006adaptive}:
\begin{align}\label{eq:penalty}
    \mathcal{P}_{\gamma}(\mathbf{M}) = \sum_{j,k=1}^d c_{jk}^{\gamma}|m_{jk}|, \quad c_{jk}>0.
\end{align}
A typical example of the weight is $c_{jk}=1/|m_{jk}^0|$, where $m_{jk}^0$ is an initial estimator for $m_{jk}$.
A candidate for $m_{jk}^0$ is the MLE or the estimate at a small penalty with $c_{jk}^{\gamma}=1$.
The tuning parameter $\gamma$ can be selected by cross-validation. However, we employ $\gamma=1$ for simplicity, which was used in some relevant papers \cite{zhang2009ica,hyvarinen2010estimation}.
Unfortunately, we cannot impose sparsity on $\mathbf{B}$ directly because the MLE cannot identify the order of ICs, and $\mathbf{B}$ cannot be derived correctly. In this paper, the order of the ICs is decided in a similar way to ICA-LiNGAM.

Here we review the principal component analysis (PCA) with sparsity. The PCs are obtained under orthogonality constraint. This situation is similar to that discussed above. 
\citet{jolliffe2003modified} proposed a method for obtaining sparse PCs by the maximizing the explained variances with an $L_1$ penalty under orthogonality constraint. \citet{zou2006sparse} reported that it was difficult to obtain sufficiently sparse PCs by the method of \citet{jolliffe2003modified}, and then they proposed {a smart idea} to obtain sparse PCs. Unfortunately, such an idea cannot be applied to our situation. The most important message is that the sparse penalty may not work well under the orthogonality constraint. To address this issue, a novel method is proposed {in this paper}, which contains an additional penalty related to the orthogonality constraint.

In order to obtain a sparse structure on $\mathbf{M}$, we relax the orthogonality constraint on $\mathbf{W}$. 
More precisely, we impose a unit-norm constraint on each row of $\mathbf{W}$ and  relax the off-diagonal orthogonality constraint by adding a new penalty term $\|\mathbf{P}^T\mathbf{W}-\mathbf{I}\|_F^2$ with $\mathbf{P}^T\mathbf{P}=\mathbf{I}$
to the objective function. Note that the additional penalty term implies the orthogonality constraint because the minimizer of $\|\mathbf{P}^T\mathbf{W}-\mathbf{I}\|_F^2$ under $\mathbf{P}^T\mathbf{P}=\mathbf{I}$ satisfies $\mathbf{W}^T\mathbf{W}=\mathbf{I}$.

Finally, we summarize the proposed method. The objective function consists of the log-likelihood \eqref{eq:likW}, the adaptive lasso penalty \eqref{eq:penalty}, and the orthogonality penalty $\|\mathbf{P}^T\mathbf{W}-I\|_F^2$ with $\mathbf{P}^T\mathbf{P}=I$. Denote the objective function by 
    \begin{gather}
    \begin{array}{l}
        F(\mathbf{W}) =
            -\tilde{\ell}(\mathbf{W}; \mathbf{X})\\
            \qquad +\lambda\left\{
                \alpha\mathcal{P}_{\gamma}(\mathbf{WD}^{-1}\mathbf{V}^T)+\frac{(1-\alpha)}{2}\|\mathbf{P}^T\mathbf{W}-\mathbf{I}\|_F^2
            \right\}.
    \end{array} \notag
    \end{gather}
    Let $\mathcal{N}\subset\mathbb{R}^{d\times d}$ be the set of non-singular matrices whose row vectors are normalized. 
    The parameter estimation is defined by
    \begin{align} \label{eq:optim}
        \begin{array}{cc}\displaystyle
            \hat{\mathbf{W}}=\min_{\mathbf{W}\in\mathcal{N}} F(\mathbf{W}) \qquad
            \text{subject to} \quad \mathbf{P}^T\mathbf{P}=\mathbf{I}. 
        \end{array}
    \end{align}
The probability density functions $\tilde{p}_j$s are adaptively selected {between} two candidates, as seen in \citet{hyvarinen2001independent}. After estimating $\mathbf{M}$ by $\hat{\mathbf{M}}=\hat{\mathbf{W}}\mathbf{D}^{-1}\mathbf{V}^T$, we recover the causal order of the variables. The estimation algorithms for the parameters and the causal order are described in the next section.

Note that each column of $\mathbf{X}$ is centered and normalized before obtaining $\mathbf{V}$ and $\mathbf{D}$, as adopted for a usual lasso. 
The tuning parameter $\lambda$ controls the total extent of the penalty, and $\alpha$ balances the sparsity and the orthogonality. 

\section{Algorithm}

In this section, we show the whole algorithm of the proposed method.
First, we show how to obtain the minimizer $\hat{\mathbf{W}}$ with given tuning parameters $\lambda$ and $\alpha$. Next, we illustrate how to select the tuning parameters. 
Finally, we explain how to estimate the causal order of variables. 

\subsection{How to Obtain Parameter Estimates}
    We derive an optimization algorithm based on the Alternating Direction Method of Multipliers \citep[ADMM;][]{boyd2011distributed}.
    ADMM is applied to the optimization of the objective function $f_1(x)+f_2(z)$ under a linear constraint on $(x,z)$. In particular, ADMM is a strong tool when one of two functions is simple, and then ADMM is widely used to solve a sparse estimation problem.
    
    First, the optimization problem \eqref{eq:optim} is equivalently transformed to the ADMM form:
    \begin{align} \label{eq:ADMM}
        \begin{array}{cc}\displaystyle
            \min_{\mathbf{W}\in\mathcal{N},\mathbf{M,P}\in\mathbb{R}^{d\times d}} 
                F_1(\mathbf{W},\mathbf{P}) + F_2(\mathbf{M})\\
            \text{subject to} \quad
                \mathbf{WD}^{-1}\mathbf{V}^T = \mathbf{M},~
                \mathbf{P}^T\mathbf{P}=\mathbf{I},
        \end{array}
    \end{align}
    where
    \begin{align}
        F_1(\mathbf{W},\mathbf{P}) 
            &= -\tilde{\ell}(\mathbf{W}; \mathbf{X})
                + \lambda\frac{(1-\alpha)}{2}\|\mathbf{P}^T\mathbf{W}-\mathbf{I}\|_F^2,\notag\\
        F_2(\mathbf{M}) 
            &= \lambda\alpha\mathcal{P}_{\gamma}(\mathbf{M}).\notag
    \end{align}    
    Then, we update $(\mathbf{W},\mathbf{M})$ and $\mathbf{P}$ alternately.
    For given $\mathbf{W}$ (and $\mathbf{M}$), the updated matrix of $\mathbf{P}$ is obtained by minimizing $\|\mathbf{P}^T\mathbf{W}-\mathbf{I}\|_F^2$ under $\mathbf{P}^T\mathbf{P}=\mathbf{I}$. Let {the singular value decomposition of $\mathbf{W}$ be denoted by $\mathbf{U}_W \mathbf{D}_W \mathbf{V}_W^T$.} We see that the updated matrix of $\mathbf{P}$ is given by $\mathbf{P} = \mathbf{U}_W\mathbf{V}_W^T$. (Mardia, Kent, and Bibby 1979).
    For given $\mathbf{P}$, the augmented Lagrangian is given by
    \begin{align}
    & \mathcal{L}_{\rho}(\mathbf{W},\mathbf{M}, \mathbf{U}) = 
            F_1(\mathbf{W},\mathbf{P})+F_2(\mathbf{M})  \label{eq:lagrange} \\
    & \ \          +\mathrm{tr}\left[\mathbf{U}^{T}(\mathbf{WD}^{-1}\mathbf{V}^T-\mathbf{M})\right]
            + \frac{\rho}{2}\|\mathbf{WD}^{-1}\mathbf{V}^T - \mathbf{M}\|_F^2, \nonumber
    \end{align} 
    where $\mathbf{U}\in\mathbb{R}^{d\times d}$ is a Lagrange multiplier matrix and $\rho$ is a tuning parameter. From the optimality condition of ADMM, 
    the updates of $\mathbf{W},\mathbf{M}, \text{and}~\mathbf{U}$ are given by
    \begin{align} \label{eq:updates}\left\{
        \begin{array}{l}
            \displaystyle \mathbf{W}_{t+1}=\mathrm{arg}\min_{\mathbf{W}\in\mathcal{N}}\mathcal{L}_{\rho}(\mathbf{W},\mathbf{M}_t,\mathbf{U}_t) \\
            \displaystyle \mathbf{M}_{t+1}
            =\mathrm{arg}\min_{\mathbf{M}\in\mathbb{R}^{d\times d}}\mathcal{L}_{\rho}(\mathbf{W}_{t+1},\mathbf{M},\mathbf{U}_t) \\
            \displaystyle \mathbf{U}_{t+1}=\mathbf{U}_t+\rho(\mathbf{W}_{t+1}\mathbf{D}^{-1}\mathbf{V}^T-\mathbf{M}_{t+1})
        \end{array}\right. ,
    \end{align}
    In the following, the first and second updates are discussed in detail. 

    \begin{algorithm}[H]
	    \caption{Gradient Descent for $\mathbf{W}$}\label{alg:updateW}
	    \begin{algorithmic}[1]
		    \State
		    \textbf{Input:} \newline 
		        $\mathbf{W}_t$,~$\mathcal{L}_{\rho}(\mathbf{W},\mathbf{M}_t,\mathbf{U}_t)$, $u_{\mathrm{max}}$, and learning rate $\eta>0$ \\
            \textbf{Output:} $\mathbf{W}_{t+1}$
            \vspace{3mm}
            \State{Calculate the natural gradient $\Delta\tilde{\mathbf{W}}$ and its modification $\Delta{\mathbf{W}}$}.
		    \For{$u = 1$ to $u_{\mathrm{max}}$}
			    \State{$\mathbf{W}_t^{(u+1)} \leftarrow \mathbf{W}_t^{(u)} - \eta\Delta\mathbf{W}$}
			    \State{{\bf break} if the convergence criteria for $\mathbf{W}$ is satisfied.}
    		\EndFor{\textbf{end for}}
			\State{Update $\mathbf{W}$ as $\mathbf{W}_{t+1}\leftarrow\mathbf{W}_{t}^{(u_{\mathrm{max}})}$}
			\State{$\boldsymbol{w}_{j,t+1} \leftarrow \boldsymbol{w}_{j,t+1}/\|\boldsymbol{w}_{j,t+1}\|$ for all $j$}
	    \end{algorithmic}
    \end{algorithm}
    The first update in \eqref{eq:updates} is based on the gradient descent in Algorithm~\ref{alg:updateW}. The key is the gradient $\Delta\mathbf{W}$. {We use} the natural gradient \cite{amari1998natural}, given by
    \begin{align}
        \Delta\tilde{\mathbf{W}}
            &= \frac{\partial\mathcal{L}_{\rho}(\mathbf{W},\mathbf{M}_t,\mathbf{U}_t)}{\partial\mathbf{W}}
                \mathbf{W}^T\mathbf{W} \notag\\
            &= -\left(\frac{1}{N}\sum_{i=1}^N\sum_{j=1}^d
                \tilde{g}(\boldsymbol{w}_j^T\boldsymbol{z}_i)\boldsymbol{w}_j^T\boldsymbol{z}_i+\mathbf{I}\right)\mathbf{W} \notag\\
                &\quad+\lambda\frac{(1-\alpha)}{2}(\mathbf{W}-\mathbf{P})\mathbf{W}^T\mathbf{W}\label{eq:natgradW}\\
                &\quad+\left\{\mathbf{U}_t+\rho(\mathbf{W}\mathbf{D}^{-1}\mathbf{V}^T-\mathbf{M}_t)
                \right\}\mathbf{VD}^{-1}\mathbf{W}^T\mathbf{W}.\notag
    \end{align}
    The problem is that the updated matrix using the natural gradient is usually not in $\mathcal{N}$, so that we must make a gradient such that the updated matrix is in $\mathcal{N}$.
    Consider a small change $\mathbf{W}+\varepsilon\Delta\mathbf{W}~(\varepsilon>0)$. When it is in $\mathcal{N}$, the diagonal entries of     $(\mathbf{W}+\varepsilon\Delta\mathbf{W})(\mathbf{W}+\varepsilon\Delta\mathbf{W})^T$ must be one. Here we ignore a very small term related to $\varepsilon^2$, which yields $\mathrm{diag}(\mathbf{W}\Delta\mathbf{W}^T)=\mathbf{0}$. 
    Let $\Delta\mathbf{W}=(\delta\boldsymbol{w}_1,\ldots,\delta\boldsymbol{w}_d)^T$ with
    \begin{align}
        \delta\boldsymbol{w}_j 
            &= \delta\tilde{\boldsymbol{w}}_j - \frac{\langle\boldsymbol{w}_j,\delta\tilde{\boldsymbol{w}}_j\rangle}{\|\boldsymbol{w}_j\|^2}\boldsymbol{w}_i
            ~~\text{for}~j=1,2,\ldots,d. 
    \end{align}
    We can easily see $\mathrm{diag}(\mathbf{W}\Delta\mathbf{W}^T)=\mathbf{0}$. Instead of the natural gradient $\Delta\tilde{\mathbf{W}}$, we use the modified gradient $\Delta\mathbf{W}$ in Algorithm~\ref{alg:updateW}. Rigorously, the resulting updated matrix is slightly out of $\mathcal{N}$. Hence, at the 8th step in Algorithm~\ref{alg:updateW}, it is pulled back to $\mathcal{N}$.

    Although the first update in \eqref{eq:updates} requires the minimization of $\mathcal{L}_{\rho}(\mathbf{W},\mathbf{M}_t,\mathbf{U}_t)$ at each step, it is computationally heavy. Instead, we set an upper limit $u_\mathrm{max}$ on the number of iterations to make the algorithm fast. From our experience, even if $u_\mathrm{max}$ is not large, the algorithm converges well.

    The second update in \eqref{eq:updates} can be expressed in a closed form. The subgradient equation of  $\mathcal{L}_{\rho}(\mathbf{W}_{t+1},\mathbf{M},\mathbf{U})$ with respect to $\mathbf{M}$ is 
    \begin{align}
        \lambda\alpha\partial_{\mathbf{M}}\mathcal{P}_{\gamma}(\mathbf{M})
            - \mathbf{\mathbf{U}}_t 
            - \rho(\mathbf{W}_{t+1}\mathbf{D}^{-1}\mathbf{V}^T - \mathbf{M}) = O.
    \end{align}
    From this equation, the update for $\mathbf{M}$ takes the form
    \begin{align}
        \mathbf{M}_{t+1} 
            = \mathcal{S}\left(\mathbf{W}_{t+1}\mathbf{D}^{-1}\mathbf{V}^T + \frac{1}{\rho}\mathbf{\mathbf{U}}_t~;~\frac{\lambda\alpha}{\rho}\mathbf{C}_\gamma\right),
    \end{align}
    where $\mathbf{C}_\gamma= (c_{jk}^\gamma)${, and} $\mathcal{S}(\cdot)$ is the soft-thresholding operator, given by
    \begin{align}
        \{\mathcal{S}(X;\mathbf{C})\}_{jk} = \left\{\begin{array}{cc}
            X_{jk} - c_{jk} & (X_{jk} > c_{jk})\\
            0 &  (- c_{jk}\le X_{jk} \le  c_{jk})\\ 
            X_{jk} + c_{jk} & (X_{jk}< -c_{jk})
        \end{array}\right..\notag
    \end{align}

    ADMM usually has one stopping criterion based on the linear constraint, but the proposed algorithm additionally requires another one because we update $\mathbf{W}$ in an iterative manner. Detailed criteria are described in the {numerical experiments}.

    \subsection{Tuning Parameter Selection}
    The tuning parameter $\lambda(1-\alpha)$ should be large because we expect that the orthogonality constraint holds approximately. The tuning parameter $\alpha$ balances the sparsity and the orthogonality. In order to achieve these purposes, $\lambda$ is set to be large, and $\alpha$ is selected in the neighborhood of zero by the K-fold CV. In this strategy, the tuning parameter for the sparsity, $\lambda\alpha$, can be taken from zero to a large value.

    Many sparse estimation methods search for the tuning parameter in descending order \cite{hastie2015statistical}, but our algorithm does in ascending order. There are two reasons.
    One reason is that it is impossible to obtain a maximum of $\lambda\alpha$. For example, in lasso \cite{tibshirani1996regression}, the maximum tuning parameter is obtained so that all the estimates are shrunk to zero. In contrast, our methods expect $\hat{\mathbf{M}}$ to be transformed to $\hat{\mathbf{B}}$, which means $\hat{\mathbf{M}}$ must have at least $d$ non-zero values. Another reason is that we often encountered that the algorithm fell into an inappropriate local minimum with a large $\lambda\alpha$.
    
    There is another remark on K-fold CV in the proposed method. We usually use the average of the log-likelihood paths over the K folds, and we choose the $\lambda$ which maximizes the averaged path. However, we observed that a log-likelihood path rarely changed more steeply than other paths. In that case, the averaged path is influenced by the irregular one. 
    For a robust selection of $\lambda$, we obtain K $\lambda$s that give the maximum value of each path and then take the median of them.

\begin{figure*}[t]
\centering
\includegraphics[width=0.92
\textwidth]{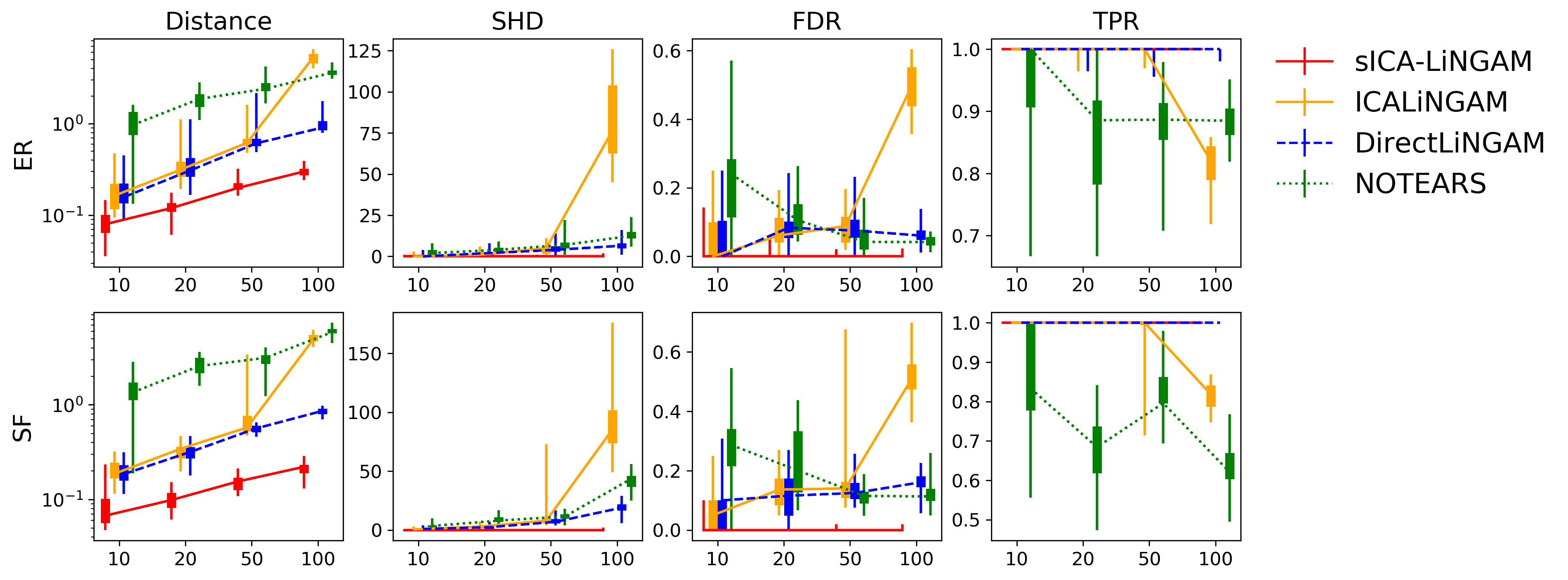} 
\caption{
Four evaluation measures (Distance, SHD, FDR, TPR) over two graph types (ER, SF) and four graph sizes (10, 20, 50, 100). The x-axis is the graph size, and the y-axis is the value of each measure. 
The main line indicates the median of 20 simulations. The thick error bar and the thin error bar are the interquartile range and min-max range, respectively.}
\label{fig_main}
\end{figure*}

    \subsection{Post-Processing}
    We obtain the estimate $\hat{\mathbf{W}}$ from the optimization problem \eqref{eq:optim}. However, we cannot estimate $\mathbf{B}$ directly via the relation $\hat{\mathbf{B}}=\mathbf{I}-\hat{\mathbf{M}}$ because of two problems, which are already described:
    (i) {r}ows of $\hat{\mathbf{M}}$ have to be rearranged and rescaled so that all diagonal entries are one{, and}
    (ii) {t}he estimate $\hat{\mathbf{B}}$ may not be acyclic, even if the true $\mathbf{B}$ is acyclic.
    These problems can be resolved by the method proposed in \citet{shimizu2006linear}. Here, we only show the outline of this method.

    The problem (i) is resolved as follows. First, in order to obtain the matrix with non-zero diagonal entries, we search for a row permutation $\pi$ minimizing $\sum_{j=1}^d 1/|\{\pi(\hat{\mathbf{M}})\}_{jj}|$. Next, we rescale the diagonal entries of $\pi(\hat{\mathbf{M}})$ to be one and divide each row by the same scale.

    The problem (ii) is resolved as follows.
    To ensure the acyclicity, we repeatedly apply a test-and-cutoff procedure:
    \begin{enumerate}
        \item Test whether $\hat{\mathbf{B}}$ is acyclic or not.
        \item If $\hat{\mathbf{B}}$ is not acyclic, replace the non-zero smallest absolute value of $\hat{\mathbf{B}}$ to 0, and return to step 1.
    \end{enumerate}
    For the test of acyclicity, an efficient algorithm was proposed by \citet{shimizu2006linear}.
    
    There is an additional device in the proposed method. After $\hat{\mathbf{B}}$ is made acyclic, the cutoff value is obtained. When this value is larger than pre-specified criteria $\omega_1(>0)$, such as $\omega_1 = 0.05$, we improve the estimate by increasing $\alpha$ and {again} estimating $\hat{\mathbf{M}}$ {until $\hat{\mathbf{B}}$ is made acyclic with a smaller cutoff than $\omega_1$}. This truncation seems ad-hoc, but relevant methods like NOTEARS set larger criteria like $\omega_1=0.3$. 
    The proposed method worked well with much smaller cutoffs in numerical experiments, {as shown in Figure~\ref{fig_cut}}.
    Furthermore, we can reduce false discovery of directed edges by additional cutoff $\omega_2>0$ for which the entries of $\hat{\mathbf{B}}$ is truncated to 0 if their absolute values are smaller than $\omega_2$.

\section{Experiments}

In this section, we show the performance of the proposed method by numerical experiments. On synthetic data, the proposed method (sICA-LiNGAM) was compared with ICA-LiNGAM, DirectLiNGAM, and NOTEARS. We did not include Pairwise LiNGAM, High-dimensional LiNGAM, and the method of \citet{zhang2009ica} because the first two methods {would show} similar performances to DirectLiNGAM in our setting, as seen in \citet{hyvarinen2013pairwise} and \citet{wang2020high}, and the last method is considered to be unstable as discussed before.
The scalability of the proposed method was examined in high dimensional data up to $d=500$ with a fixed sample size. The application to the real data was also conducted.

\subsection{Comparison of the methods (by simulation)}
Numerical experiments were conducted on synthetic data. 
We basically mimic the simulation settings of \citet{shimizu2011directlingam} and \citet{zheng2018dags}.
The linear DAG model with non-Gaussian noises was employed to generate observed variables. True graphs were generated from {\it Erd\"{o}s-R\'{e}nyi} {\cite[ER;][]{erdos1959random}} or {\it Scale-Free} {\cite[SF;][]{barabasi1999emergence}} models of different sizes ($d=10,20,50,100$), with the expected number of edges $d$.
The weight parameters of a generated graph were uniformly drawn from the interval $[-1.5,-0.5]\cup [0.5,1.5]$. (The true non-zero weight parameters were avoided to be around zero.)
The distribution of each noise {was} randomly selected from three non-Gaussian distributions ({\it Laplace, uniform}, and {\it exponential}). The noise variances were uniformly drawn from the interval $[1,3]$. We generated 20 datasets with a sample size of $N=1,000$. 
The noises were randomly drawn from the selected distributions.

The models were estimated by the proposed method, ICA-LiNGAM, DirectLiNGAM, and NOTEARS. 
We describe the settings for these methods. 
For the proposed method, the tuning parameter $\lambda$ {was} set to $\lambda = 0.1$ throughout the experiments. The tuning parameter $\alpha$ {was} basically selected by 10-fold CV out of 50 points in {the} range $[10^{-3},10^{-0.5}]$ with a log scale. The initial estimate $(m_{jk}^{0})$ was estimated by the proposed method with $\gamma=0$, $\alpha=0$ for $d<50$, and $\alpha=0.1$ for $d\ge 50$. The step size was set at $\eta = 0.005$.
If the resulting estimate $\hat{\mathbf{B}}$ { was} not acyclic, the tuning parameter $\alpha$ increased until $\hat{\mathbf{B}}$ became acyclic. The two cutoff parameters $\omega_1$ and $\omega_2$ were both set at 0.05. Two stopping criteria for parameter updates were set to $\max{|\mathbf{W}_t\mathbf{D}^{-1}\mathbf{V}^T-\mathbf{M}_t|}<10^{-4}$ and $\max{|\mathbf{W}_{t+1}-\mathbf{W}_t|}<10^{-5}$. The proposed method was implemented in Python 3.6.8. For other methods, the authors' implementations were used 
\footnote[1]{ICA- and DirectLiNGAM: https://github.com/cdt15/lingam, NOTEARS: https://github.com/xunzheng/notears}.

The methods were evaluated in terms of estimation error based on the Frobenius norm between the estimated and true weight matrices (Distance), Structural Hamming Distance (SHD), False Discovery Rate (FDR) and True Positive Rate (TPR). The metrics are detailed in the supplementary material. Figure~\ref{fig_main} shows the main results. The proposed method achieved the best performances among all of the methods. This result was probably because the proposed method estimates the causal order and parameters simultaneously via {the} log-likelihood optimization with sparsity. This means the proposed method used all the information throughout the estimation process. The proposed method succeeded in improving ICA-LiNGAM by virtue of sparsity and, in particular, largely improved in the high dimensional setting $(d=100)$. NOTEARS was behind the others. It would be because the synthetic data did not satisfy the equal variance assumption, which seems necessary for NOTEARS.

\begin{figure}[H]
\centering
\includegraphics[width=0.35\textwidth]{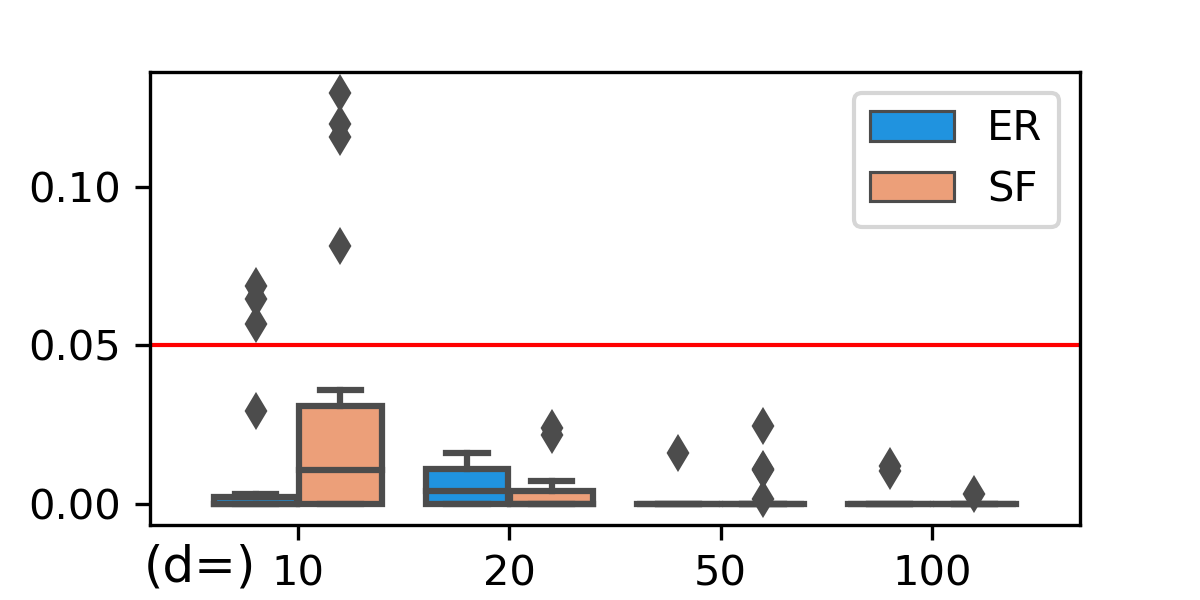} 
\caption{Cutoff values necessary to make $\hat{\mathbf{B}}$ acyclic at CV-selected $\alpha$. If the cutoff exceeded the criteria (red line), we increased $\alpha$.}
\label{fig_cut}
\end{figure}

Figure~\ref{fig_cut} shows the actual cutoff values necessary to make $\hat{\mathbf{B}}$ acyclic at CV-selected $\alpha$. Although we used $\omega_1 = 0.05$, the estimate $\hat{\mathbf{B}}$ could be acyclic with much smaller cutoff values in most cases. This indicates that we can estimate the acyclic graph in a much less ad-hoc manner.

\subsection{Scalability of the proposed method}

The scalability of the proposed method was examined in simpler and higher dimensional settings. The graphs {were generated from the ER model} with $d=100,200,500$, and noises {were} drawn only from {the} {\it Laplace} distribution. The expected number of directed edges was 50 or 100. The sample size and the other settings were the same as before. We evaluated the estimation error and computational time of the proposed method and DirectLiNGAM because these methods performed well {for $d=100$} in the previous experiment. In order to evaluate the computational time conveniently, $\alpha$ of the proposed methods was fixed at 0.1. The experiments were conducted with a single 3.6 GHz Intel Core i7 CPU and a 32GB memory.

\begin{figure}[H]
\centering
\includegraphics[width=0.4\textwidth]{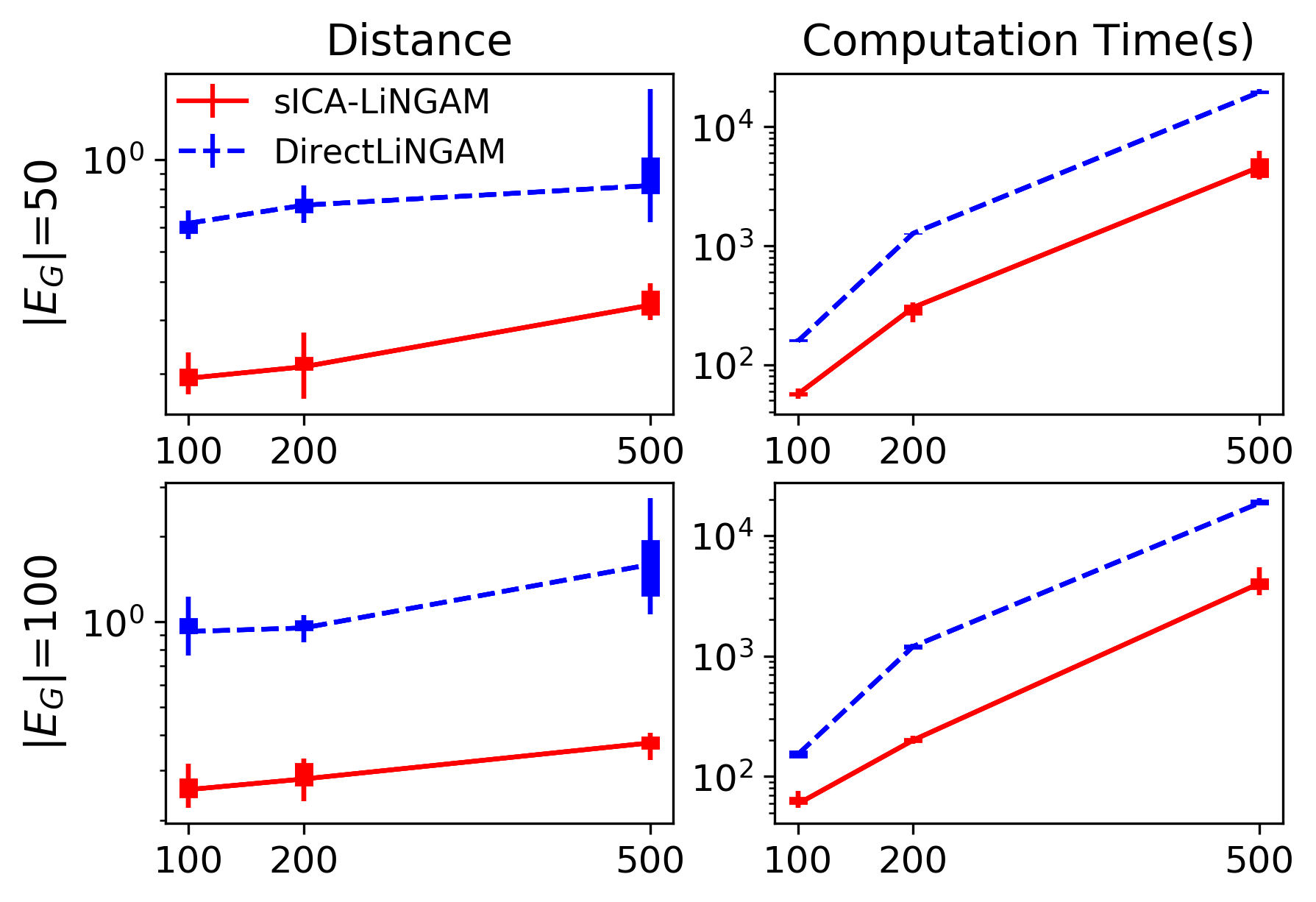} 
\caption{
Estimation error and computation time based on 10 simulations. The x-axis is the  dimension of the data. 
}
\label{fig_scale}
\end{figure}

The results of 10 simulations are shown in Figure \ref{fig_scale}. Our method performed well, even in high dimensional settings. On the contrary, the estimate of DirectLiNGAM was unstable when $d=500$. Both methods finished in realistic computational times. 

\subsection{Real Data}

As described in \citet{shimizu2006linear}, a time series can be approximated by a linear DAG model, especially if the time series is a stationary AR(1). For example, when a time series is sliced into time windows with three time points, the AR(1) structure can be approximated by the linear DAG model of
\begin{align}
    \left\{\begin{array}{cl}
        X_1 &= \varepsilon_1 \\
        X_2 &= b_{21}X_1 + \varepsilon_2 \\
        X_3 &= b_{32}X_2 + \varepsilon_3 
    \end{array}\right..
\end{align}
Therefore, if the noises are independent and non-Gaussian, LiNGAM can express the AR(1) structure, and then recover the correct order from sliced time series.

We applied the same four methods to the Beijing Multi-Site Air-Quality Data \cite{zhang2017cautionary}. This data includes hourly concentration measures of major air pollutants, such as nitrogen dioxide ($\mathrm{NO}_2$) and sulfur dioxide ($\mathrm{SO}_2$), recorded {at} national monitoring sites in Beijing. Each site provides every pollutant's time series from March 1st, 2013 to February 28th, 2017. Each time series was sliced into 1,461 time windows with 24 time points so that each window consists of hourly measures of one day. The values were transformed by the function $\log(1+x)$. When a time window contained missing values, the window was removed. Every method was evaluated by a heatmap visualizing $\hat{\mathbf{B}}$.
Suppose the AR(1) structure is recovered by the linear DAG model, only the $(j+1,j)$th cell is colored for $j=1,...,23$, and the other cells are not colored. Note that the proposed method used the CV-selected $\alpha$ because it was difficult to obtain the  acyclic estimates by increasing $\alpha$.

\begin{figure}[H]
\centering
\includegraphics[width=0.4\textwidth]{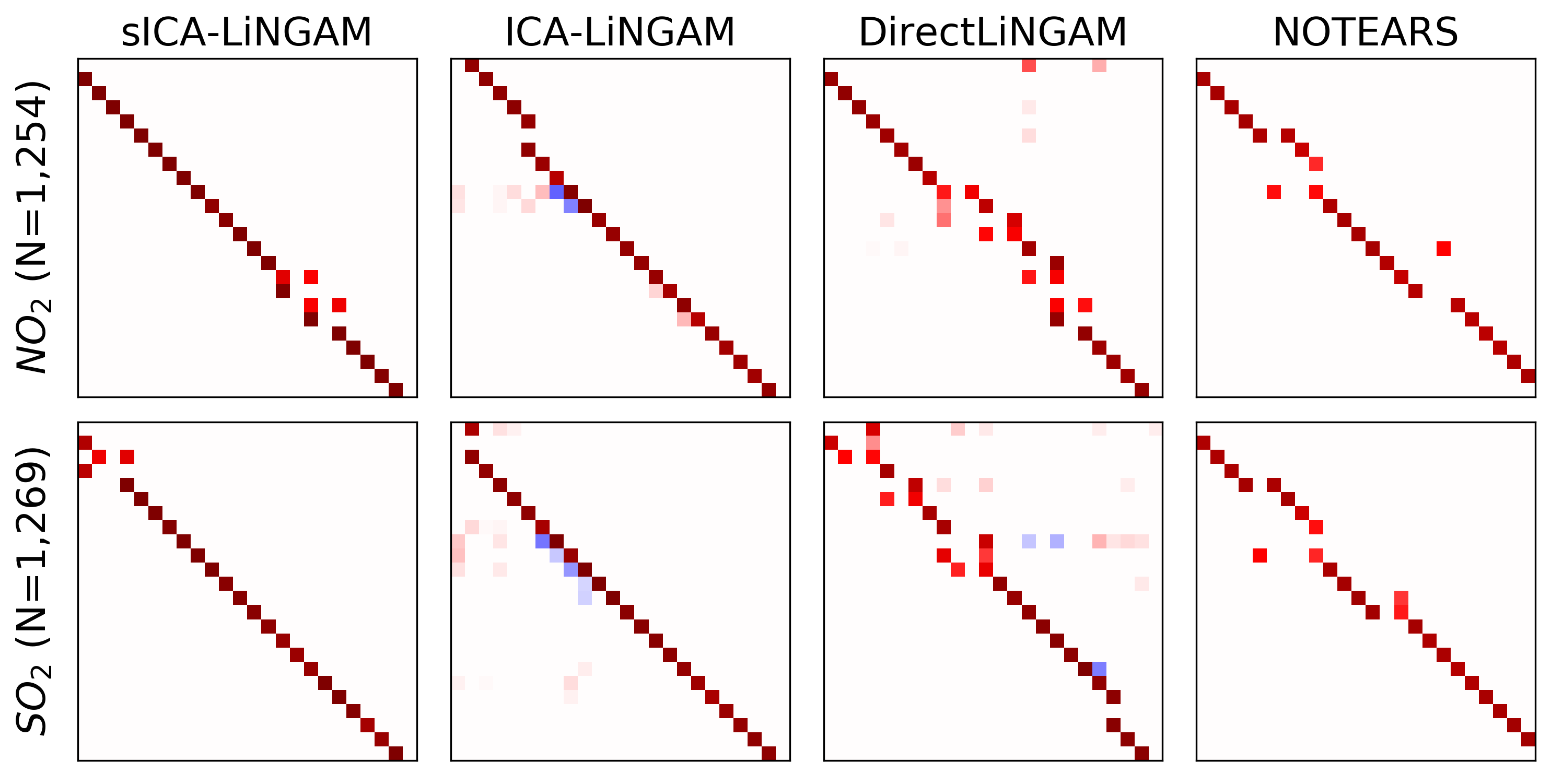}
\caption{Heatmaps of $\hat{\mathbf{B}}$s at Tiantan station. Red/Blue indicates a positive/negative value. If the AR(1) structure is recovered correctly, the $(j+1,j)$th cell is colored for $j=1,...,23$ and the other cells are not colored.}
\label{fig_heat}
\end{figure}

Figure~\ref{fig_heat} shows the results on the data of Tiantan station. The proposed method succeeded in recovering the AR(1) structure very well, and non-AR(1) cells were almost shrunk to zero.  ICA-LiNGAM also recovered the structure, but some non-AR(1) cells had non-zero values. By virtue of sparsity, the proposed method shrunk the small non-zero estimates of non-AR(1) cells to zero, and recovered the AR(1) structure better than ICA-LiNGAM. DirectLiNGAM and NOTEARS failed to recover about or more than half of the AR(1) structure. As seen in the supplementary material, the proposed method also showed the best performance clearly at the other monitoring sites.

\section{Conclusion}

In this paper, we have proposed a new estimating algorithm for a linear DAG model with non-Gaussian noises. The proposed method is based on the penalized log-likelihood of ICA, and estimates the causal structure and the parameter values at once. Several devices for a stable and efficient learning are introduced such as a penalty on orthogonality of the parameter matrix and a modified natural gradient. The proposed method achieved the best performance among the existing methods in the numerical experiments. Note that the norm constraints on the rows of $\mathbf{W}$ may not look necessary, but in our experience, the method without the constraint sometimes yielded an incorrect solution. For future works, it is significant to extend the method to non-DAG structures, such as data with cyclic structures and/or latent confounders.

\bibliography{ref}

\begin{thebibliography}{34}
\providecommand{\natexlab}[1]{#1}
\providecommand{\url}[1]{\texttt{#1}}
\providecommand{\urlprefix}{URL }
\expandafter\ifx\csname urlstyle\endcsname\relax
  \providecommand{\doi}[1]{doi:\discretionary{}{}{}#1}\else
  \providecommand{\doi}{doi:\discretionary{}{}{}\begingroup
  \urlstyle{rm}\Url}\fi

\bibitem[{Amari(1998)}]{amari1998natural}
Amari, S.-I. 1998.
\newblock Natural Gradient Works Efficiently in Learning.
\newblock \emph{Neural Computation} 10(2): 251--276.

\bibitem[{Barab{\'a}si and Albert(1999)}]{barabasi1999emergence}
Barab{\'a}si, A.-L.; and Albert, R. 1999.
\newblock Emergence of Scaling in Random Networks.
\newblock \emph{Science} 286(5439): 509--512.

\bibitem[{Bollen(1989)}]{bollen1989}
Bollen, K.~A. 1989.
\newblock \emph{Structural Equations with Latent Variables}.
\newblock John Wiley \& Sons.

\bibitem[{Boyd, Parikh, and Chu(2011)}]{boyd2011distributed}
Boyd, S.; Parikh, N.; and Chu, E. 2011.
\newblock \emph{Distributed Optimization and Statistical Learning via the
  Alternating Direction Method of Multipliers}.
\newblock Now Publishers Inc.

\bibitem[{Chickering(2002)}]{chickering2002optimal}
Chickering, D.~M. 2002.
\newblock Optimal Structure Identification with Greedy Search.
\newblock \emph{Journal of Machine Learning Research} 3(Nov): 507--554.

\bibitem[{Erd\"{o}s and R\'{e}nyi(1959)}]{erdos1959random}
Erd\"{o}s, P.; and R\'{e}nyi, A. 1959.
\newblock On Random Graphs I.
\newblock \emph{Publ. Math. Debrecen} 6(290-297): 18.

\bibitem[{Glymour(1998)}]{glymour1998learning}
Glymour, C. 1998.
\newblock Learning Causes: Psychological Explanations of Causal Explanation.
\newblock \emph{Minds and Machines} 8(1): 39--60.

\bibitem[{Hastie, Tibshirani, and Wainwright(2015)}]{hastie2015statistical}
Hastie, T.; Tibshirani, R.; and Wainwright, M. 2015.
\newblock \emph{Statistical Learning with Sparsity: the Lasso and
  Generalizations}.
\newblock CRC press.

\bibitem[{Hoover(2008)}]{hoover2008causality}
Hoover, K.~D. 2008.
\newblock Causality in Economics and Econometrics.
\newblock \emph{The new Palgrave Dictionary of Economics} 2.

\bibitem[{Hyv{\"a}rinen(1999)}]{hyvarinen1999fast}
Hyv{\"a}rinen, A. 1999.
\newblock Fast and Robust Fixed-Point Algorithms for Independent Component
  Analysis.
\newblock \emph{IEEE Transactions on Neural Networks} 10(3): 626--634.

\bibitem[{Hyv{\"a}rinen, Karhunen, and Oja(2001)}]{hyvarinen2001independent}
Hyv{\"a}rinen, A.; Karhunen, J.; and Oja, E. 2001.
\newblock \emph{Independent Component Analysis}.
\newblock John Wiley \& Sons.

\bibitem[{Hyv{\"a}rinen and Smith(2013)}]{hyvarinen2013pairwise}
Hyv{\"a}rinen, A.; and Smith, S.~M. 2013.
\newblock Pairwise Likelihood Ratios for Estimation of Non-Gaussian Structural
  Equation Models.
\newblock \emph{Journal of Machine Learning Research} 14(Jan): 111--152.

\bibitem[{Hyv{\"a}rinen et~al.(2010)Hyv{\"a}rinen, Zhang, Shimizu, and
  Hoyer}]{hyvarinen2010estimation}
Hyv{\"a}rinen, A.; Zhang, K.; Shimizu, S.; and Hoyer, P.~O. 2010.
\newblock Estimation of a Structural Vector Autoregression Model Using
  Non-Gaussianity.
\newblock \emph{Journal of Machine Learning Research} 11(5).

\bibitem[{Jolliffe, Trendafilov, and Uddin(2003)}]{jolliffe2003modified}
Jolliffe, I.~T.; Trendafilov, N.~T.; and Uddin, M. 2003.
\newblock A Modified Principal Component Technique Based on the LASSO.
\newblock \emph{Journal of Computational and Graphical Statistics} 12(3):
  531--547.

\bibitem[{Judea(2000)}]{judea2000causality}
Judea, P. 2000.
\newblock Causality: Models, Reasoning, and Inference.
\newblock \emph{Cambridge University Press. ISBN 0} 521(77362): 8.

\bibitem[{Liu et~al.(2015)Liu, Wu, Zhang, Guo, Long, and Yao}]{liu2015altered}
Liu, Y.; Wu, X.; Zhang, J.; Guo, X.; Long, Z.; and Yao, L. 2015.
\newblock Altered Effective Connectivity Model in the Default Mode Network
  Between Bipolar and Unipolar Depression Based on Resting-State fMRI.
\newblock \emph{Journal of Affective Disorders} 182: 8--17.

\bibitem[{Peters and B{\"u}hlmann(2014)}]{peters2014identifiability}
Peters, J.; and B{\"u}hlmann, P. 2014.
\newblock Identifiability of Gaussian Structural Equation Models with Equal
  Error Variances.
\newblock \emph{Biometrika} 101(1): 219--228.

\bibitem[{Peters et~al.(2014)Peters, Mooij, Janzing, and
  Sch{\"o}lkopf}]{peters2014causal}
Peters, J.; Mooij, J.~M.; Janzing, D.; and Sch{\"o}lkopf, B. 2014.
\newblock Causal Discovery with Continuous Additive Noise Models.
\newblock \emph{The Journal of Machine Learning Research} 15(1): 2009--2053.

\bibitem[{Sachs et~al.(2005)Sachs, Perez, Pe'er, Lauffenburger, and
  Nolan}]{sachs2005causal}
Sachs, K.; Perez, O.; Pe'er, D.; Lauffenburger, D.~A.; and Nolan, G.~P. 2005.
\newblock Causal Protein-Signaling Networks Derived from Multiparameter
  Single-Cell Data.
\newblock \emph{Science} 308(5721): 523--529.

\bibitem[{Shimizu et~al.(2006)Shimizu, Hoyer, Hyv{\"a}rinen, and
  Kerminen}]{shimizu2006linear}
Shimizu, S.; Hoyer, P.~O.; Hyv{\"a}rinen, A.; and Kerminen, A. 2006.
\newblock A Linear Non-Gaussian Acyclic Model for Causal Discovery.
\newblock \emph{Journal of Machine Learning Research} 7(Oct): 2003--2030.

\bibitem[{Shimizu et~al.(2011)Shimizu, Inazumi, Sogawa, Hyv{\"a}rinen,
  Kawahara, Washio, Hoyer, and Bollen}]{shimizu2011directlingam}
Shimizu, S.; Inazumi, T.; Sogawa, Y.; Hyv{\"a}rinen, A.; Kawahara, Y.; Washio,
  T.; Hoyer, P.~O.; and Bollen, K. 2011.
\newblock DirectLiNGAM: A Direct Method for Learning a Linear Non-Gaussian
  Structural Equation Model.
\newblock \emph{The Journal of Machine Learning Research} 12: 1225--1248.

\bibitem[{Spirtes et~al.(2000)Spirtes, Glymour, Scheines, and
  Heckerman}]{spirtes2000causation}
Spirtes, P.; Glymour, C.~N.; Scheines, R.; and Heckerman, D. 2000.
\newblock \emph{Causation, Prediction, and Search}.
\newblock MIT press.

\bibitem[{Tibshirani(1996)}]{tibshirani1996regression}
Tibshirani, R. 1996.
\newblock Regression Shrinkage and Selection via the Lasso.
\newblock \emph{Journal of the Royal Statistical Society: Series B
  (Methodological)} 58(1): 267--288.

\bibitem[{von Eye and DeShon(2012)}]{von2012directional}
von Eye, A.; and DeShon, R.~P. 2012.
\newblock Directional Dependence in Developmental Research.
\newblock \emph{International Journal of Behavioral Development} 36(4):
  303--312.

\bibitem[{Wang and Drton(2020)}]{wang2020high}
Wang, Y.~S.; and Drton, M. 2020.
\newblock High-Dimensional Causal Discovery Under non-Gaussianity.
\newblock \emph{Biometrika} 107(1): 41--59.

\bibitem[{Wiedermann and Von~Eye(2016)}]{wiedermann2016statistics}
Wiedermann, W.; and Von~Eye, A. 2016.
\newblock \emph{Statistics and Causality}.
\newblock John Wiley \& Sons.

\bibitem[{Xu(2017)}]{xu2017contemporaneous}
Xu, X. 2017.
\newblock Contemporaneous Causal Orderings of US Corn Cash Prices Through
  Directed Acyclic Graphs.
\newblock \emph{Empirical Economics} 52(2): 731--758.

\bibitem[{Zhang and Chan(2006)}]{zhang2006ica}
Zhang, K.; and Chan, L.-W. 2006.
\newblock ICA with Sparse Connections.
\newblock In \emph{International Conference on Intelligent Data Engineering and
  Automated Learning}, 530--537. Springer.

\bibitem[{Zhang and Hyv{\"a}rinen(2009)}]{zhang2009identifiability}
Zhang, K.; and Hyv{\"a}rinen, A. 2009.
\newblock On the Identifiability of the Post-Nonlinear Causal Model.
\newblock In \emph{25th Conference on Uncertainty in Artificial Intelligence
  (UAI 2009)}, 647--655. AUAI Press.

\bibitem[{Zhang et~al.(2009)Zhang, Peng, Chan, and
  Hyv{\"a}rinen}]{zhang2009ica}
Zhang, K.; Peng, H.; Chan, L.; and Hyv{\"a}rinen, A. 2009.
\newblock ICA with Sparse Connections: Revisited.
\newblock In \emph{International Conference on Independent Component Analysis
  and Signal Separation}, 195--202. Springer.

\bibitem[{Zhang et~al.(2017)Zhang, Guo, Dong, He, Xu, and
  Chen}]{zhang2017cautionary}
Zhang, S.; Guo, B.; Dong, A.; He, J.; Xu, Z.; and Chen, S.~X. 2017.
\newblock Cautionary Tales on Air-Quality Improvement in Beijing.
\newblock \emph{Proceedings of the Royal Society A: Mathematical, Physical and
  Engineering Sciences} 473(2205): 20170457.

\bibitem[{Zheng et~al.(2018)Zheng, Aragam, Ravikumar, and Xing}]{zheng2018dags}
Zheng, X.; Aragam, B.; Ravikumar, P.~K.; and Xing, E.~P. 2018.
\newblock DAGs with NO TEARS: Continuous Optimization for Structure Learning.
\newblock In \emph{Advances in Neural Information Processing Systems},
  9472--9483.

\bibitem[{Zou(2006)}]{zou2006adaptive}
Zou, H. 2006.
\newblock The Adaptive Lasso and Its Oracle Properties.
\newblock \emph{Journal of the American Statistical Association} 101(476):
  1418--1429.

\bibitem[{Zou, Hastie, and Tibshirani(2006)}]{zou2006sparse}
Zou, H.; Hastie, T.; and Tibshirani, R. 2006.
\newblock Sparse Principal Component Analysis.
\newblock \emph{Journal of Computational and Graphical Statistics} 15(2):
  265--286.

\end{thebibliography}

\newpage
\section{Supplementary Information}
\subsection{Definitions of Evaluation Measures}
We evaluated the estimates $\hat{\mathbf{B}}$ and the corresponding graphs by four metrics: 
1) {\it Distance}, 2) {\it Structural Hamming Distance} (SHD), 3) {\it False Discovery Rate} (FDR), 4) {\it True Positive Rate} (TPR). 
The Distance is a measure for the estimation error. The other three measures are employed to evaluate the performance of causal discovery.
\begin{itemize}
    \item {\it Distance (Frobenius Norm of the Difference Between Two Matrices)} : 
        Distance is defined as the Frobenius norm of the difference between two matrices.
        The Distance between the estimate and the truth is evaluated:
        \begin{align}
            Distance = \|\mathbf{\hat{\mathbf{B}}-\mathbf{B}_{true}}\|_F.
        \end{align}
    \item {\it Structured Hamming Distance} (SHD):
        SHD indicates the number of steps to transform the estimated graph into the true graph.
        The steps include edge additions, deletions, and reversals.
    \item {\it False Discovery Rate} (FDR): 
        FDR is the proportion of false positives and reversed edges over the estimated edges. 
    \item {\it True Positive Rate} (TPR): 
        TPR is the proportion of true positive edges over the true edges. 
\end{itemize}

\section{Results of AR(1) Recovery at the Other Sites}
    We estimated $\hat{\mathbf{B}}$ by four methods (sICA-LiNGAM (proposed), ICA-LiNGAM, DirectLiNGAM, NOTEARS) at the three additional sites. If the AR(1) structure is recovered by the linear DAG model, only the $(j+1,j)$th cell is colored for $j=1,...,23$, and the other cells are not colored.
    
    \begin{itemize}
        \item At Aotizhongxin station (Figure \ref{fig_aoti}), the proposed method successfully recovered the time series especially on $\mathrm{SO}_2$. ICA-LiNGAM partly recovered the structure, but some non-AR(1) cells had non-zero values.
        DirectLiNGAM also recovered the structure on $\mathrm{NO}_2$,
        but failed to recover about two third of the AR(1) structure on $\mathrm{SO}_2$. NOTEARS failed to recover about or more than half of the AR(1) structure.
        \item At Nongzhanguan station (Figure \ref{fig_nong}), the proposed method completely recovered the AR(1) structure except for the points of (1,24) and (21,22) on $\mathrm{SO}_2$. 
        None of the other methods could recovered the AR(1) sequence successfully.
    \end{itemize}
    
    \newpage
    
    \renewcommand{\thefigure}{S.1}
    \begin{figure}[H]
    \centering
    \includegraphics[width=0.45\textwidth]{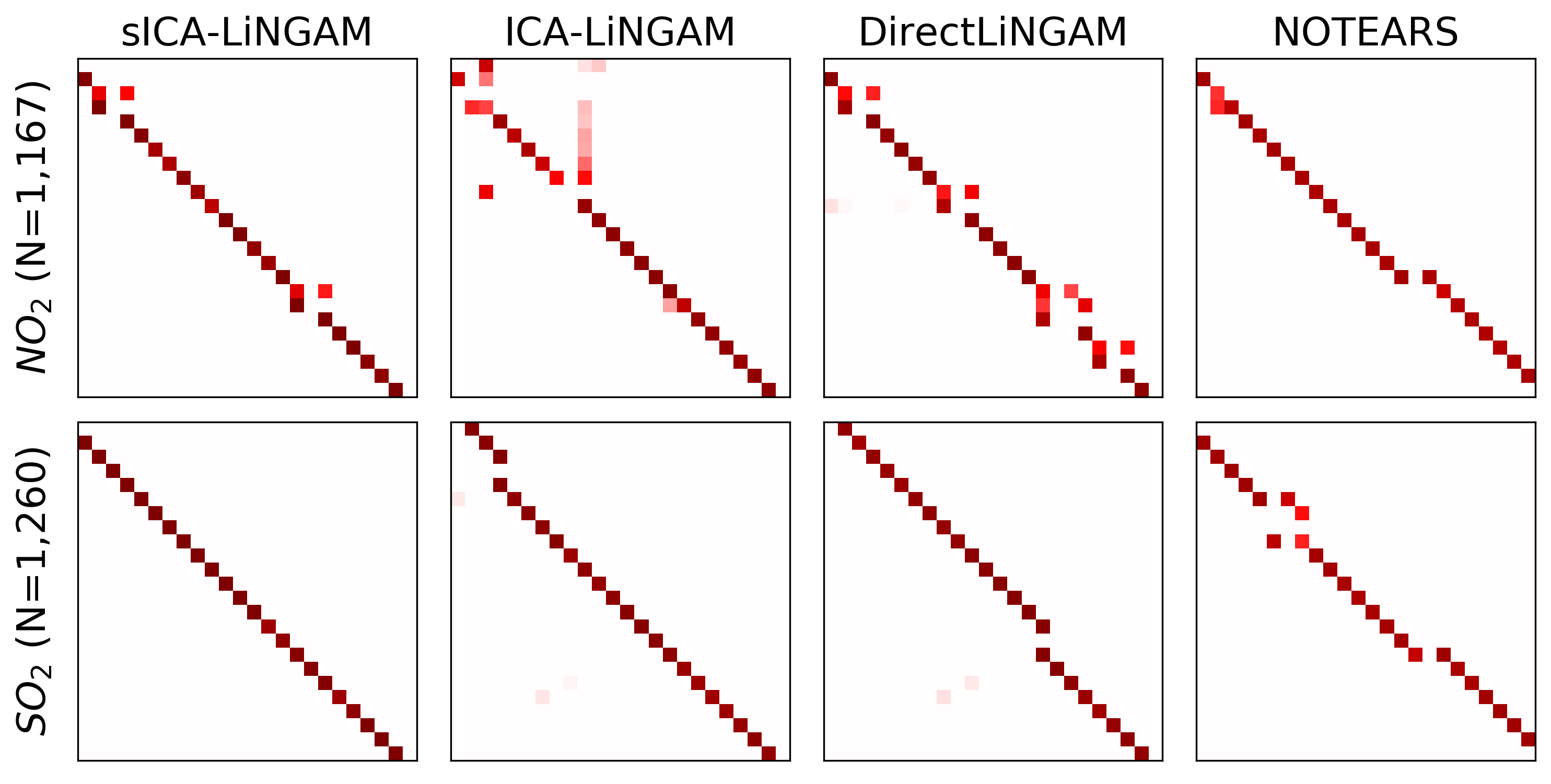}
    \caption{Heatmaps of $\hat{\mathbf{B}}$s at Aotizhongxin station. Red/Blue indicates a positive/negative value.}
    \label{fig_aoti}
    \end{figure}
    
    \renewcommand{\thefigure}{S.2}
    \begin{figure}[H]
    \centering
    \includegraphics[width=0.45\textwidth]{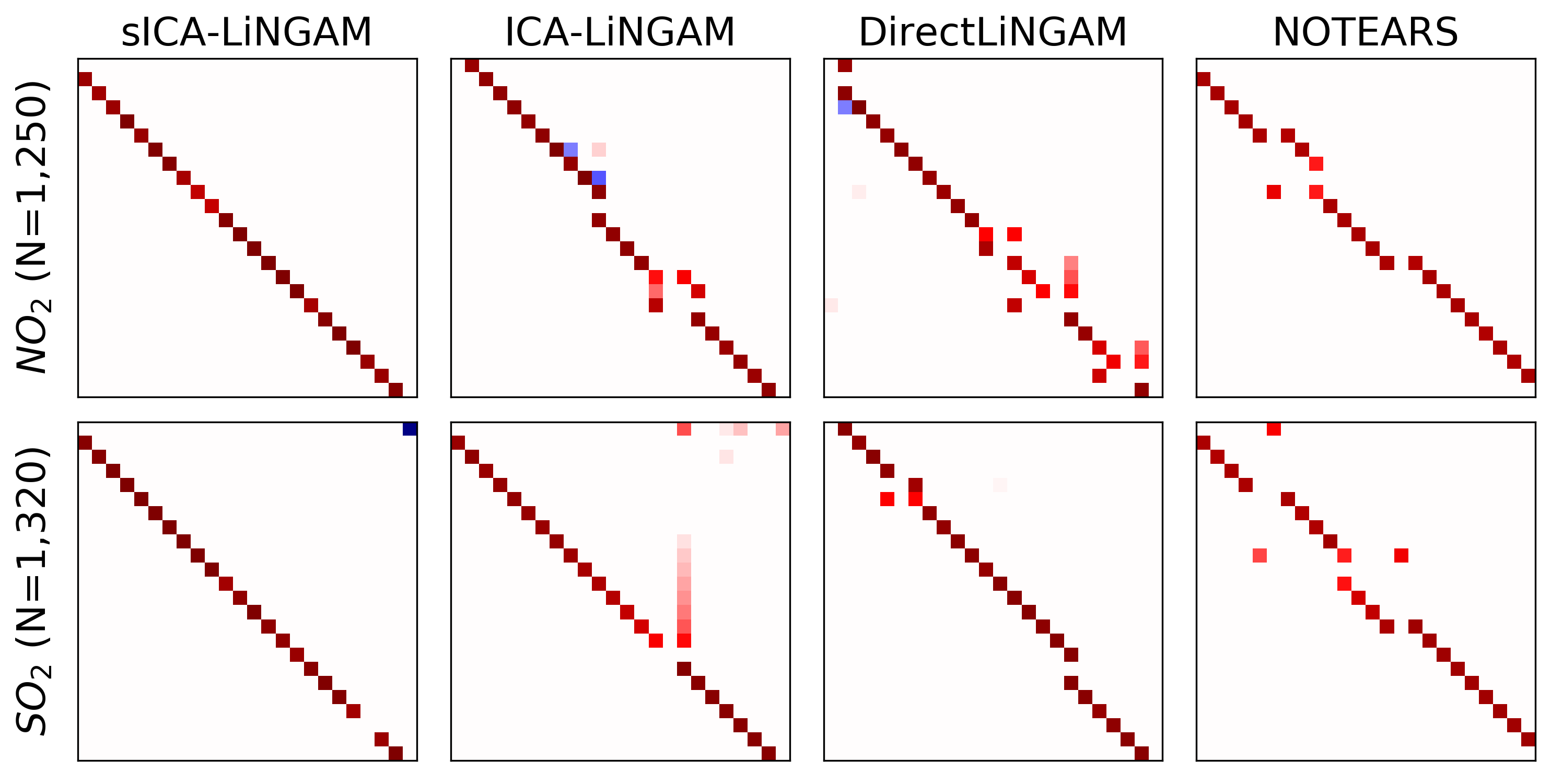}
    \caption{Heatmaps of $\hat{\mathbf{B}}$s at Nongzhanguan station. Red/Blue indicates a positive/negative value.}
    \label{fig_nong}
    \end{figure}

\end{document}